\documentclass[letterpaper, 10 pt, conference]{ieeeconf}

\IEEEoverridecommandlockouts                              
\usepackage[utf8]{inputenc}
\usepackage{comment}
\usepackage{booktabs}

\usepackage{graphicx}
\usepackage{caption}
\usepackage{subcaption}
\usepackage{amsmath}
\usepackage{verbatim}
\usepackage{amsfonts}
\usepackage{multirow}
\usepackage{changepage}
\usepackage{tabularx}
\usepackage{adjustbox}
\usepackage{mathtools}
\usepackage{amssymb}
\usepackage{soul}
\usepackage{color}
\usepackage[noend]{algpseudocode}

\usepackage{color}
\usepackage{tikz}
\usepackage{flushend}
\usetikzlibrary{shapes.geometric, arrows}
\usepackage[linesnumbered,ruled]{algorithm2e}
\usepackage{multirow}
\usepackage{soul,color}
\usepackage{graphicx}
\usepackage[colorlinks]{hyperref}
\usepackage{fancyhdr}
\usepackage{comment}
\usepackage{epsfig}
\usepackage{bm}
\usepackage{float}
\usepackage{cuted}
\usepackage{tcolorbox}
\usepackage{xcolor}
\usepackage{mdframed}
\allowdisplaybreaks

\makeatletter
\def\algbackskip{\hskip-\ALG@thistlm}
\makeatother

\newcommand{\union}{\operatorname{\cup}}
\newcommand{\intersection}{\ensuremath{\operatorname{\cap}}}

\newcommand\oprocendsymbol{\hbox{$\square$}}
\newcommand\oprocend{\relax\ifmmode\else\unskip\hfill\fi\oprocendsymbol}

\usepackage{cite}

\newcommand\red[1]{{\color{red} #1}}
\newcommand\blue[1]{{\color{blue} #1}}

\newcommand\orange[1]{{\color{orange} #1}}

\renewcommand{\theenumi}{(\roman{enumi}}

\mdfdefinestyle{custombox}{%
    backgroundcolor=gray!10, 
    linewidth=1pt,           
    linecolor=black,         
    innertopmargin=10pt,     
    innerbottommargin=10pt,  
    innerrightmargin=10pt,   
    innerleftmargin=10pt     
}

\makeatletter
\let\NAT@parse\undefined
\makeatother

\renewcommand{\baselinestretch}{0.94}

\title{\LARGE \bf
Generalized Mission Planning for Heterogeneous Multi-Robot Teams via LLM-constructed Hierarchical Trees
}

\author{
Piyush Gupta\textsuperscript{1*} \hspace{1.2cm}  David Isele\textsuperscript{1} \hspace{1.2cm} Enna Sachdeva\textsuperscript{1} \hspace{1.2cm}  Pin-Hao Huang\textsuperscript{1} \\ Behzad Dariush\textsuperscript{1} \hspace{1.2cm}  Kwonjoon Lee\textsuperscript{1} \hspace{1.2cm} 
Sangjae Bae\textsuperscript{1} 
\thanks{
\textsuperscript{1} Honda Research Institute USA, San Jose, CA, 95134, USA.}
\thanks{
\textsuperscript{*} Corresponding author \texttt{\{piyush\_gupta\}@honda-ri.com } \
}
}

\begin{document}

\maketitle
\thispagestyle{plain}
\pagestyle{plain}
\pagenumbering{gobble}
\begin{abstract}
We present a novel mission-planning strategy for heterogeneous multi-robot teams, taking into account the specific constraints and capabilities of each robot. Our approach employs hierarchical trees to systematically break down complex missions into manageable sub-tasks. We develop specialized APIs and tools, which are utilized by Large Language Models (LLMs) to efficiently construct these hierarchical trees. Once the hierarchical tree is generated, it is further decomposed to create optimized schedules for each robot, ensuring adherence to their individual constraints and capabilities. We demonstrate the effectiveness of our framework through detailed examples covering a wide range of missions, showcasing its flexibility and scalability. 

\end{abstract}


\section{Introduction}

\begin{figure}[ht]
    \centering
	\begin{subfigure}[b]{0.18\textwidth}
	    \centering
        \includegraphics[width=1\linewidth, height=1\linewidth, keepaspectratio]{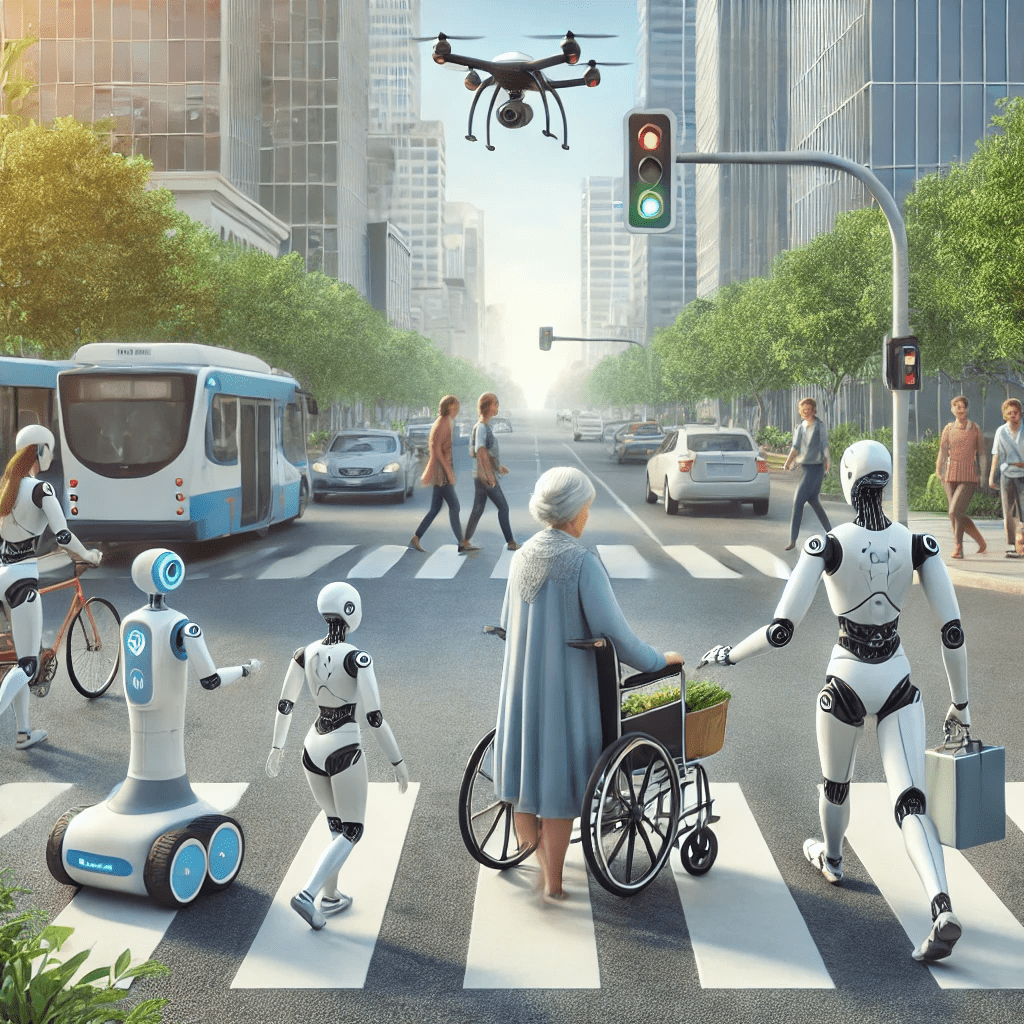}
        \caption{}
        \label{fig:oldlady}
    \end{subfigure}
    \begin{subfigure}[b]{0.18\textwidth}
	    \centering
        \includegraphics[width=1\linewidth, height=1\linewidth, keepaspectratio]{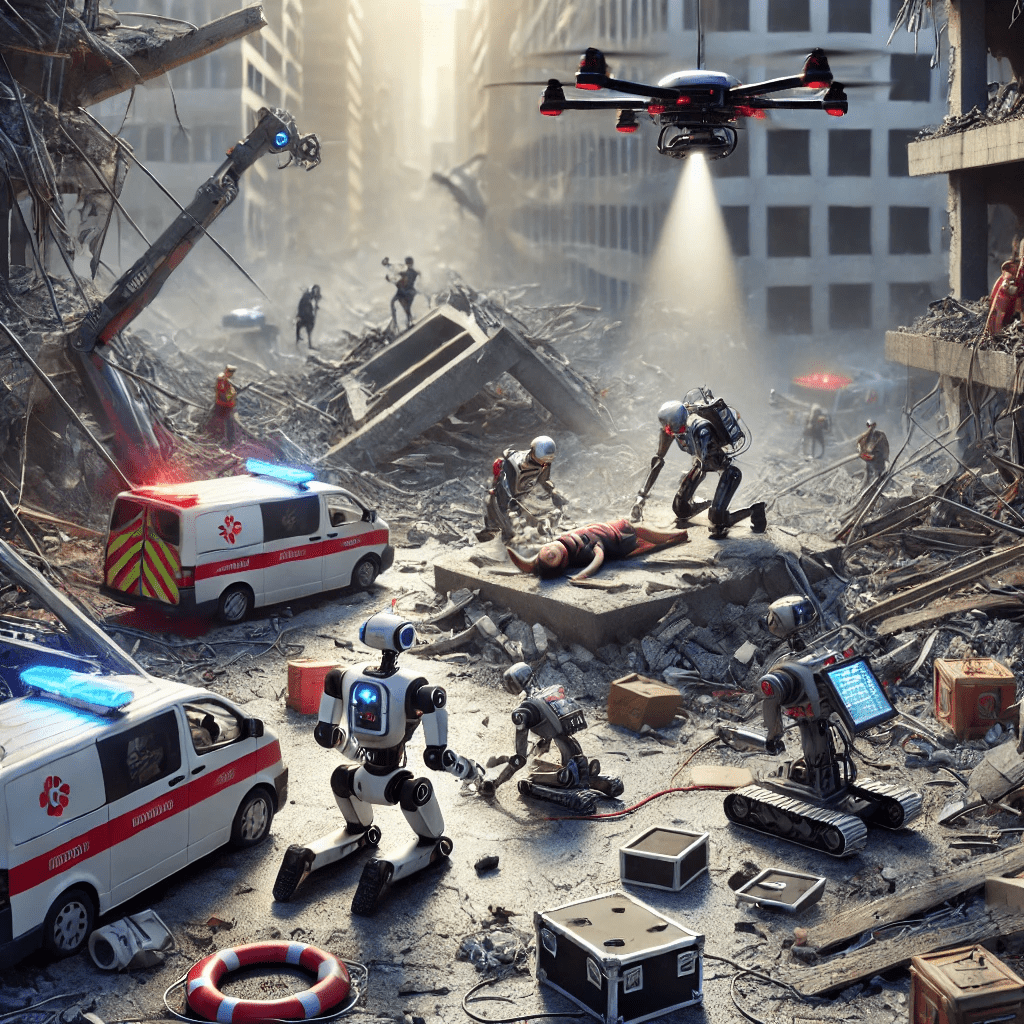}
        \caption{}
        \label{fig:searchAndRescue}
    \end{subfigure}
    	\begin{subfigure}[b]{0.18\textwidth}
	    \centering
        \includegraphics[width=1\linewidth, height=1\linewidth, keepaspectratio]{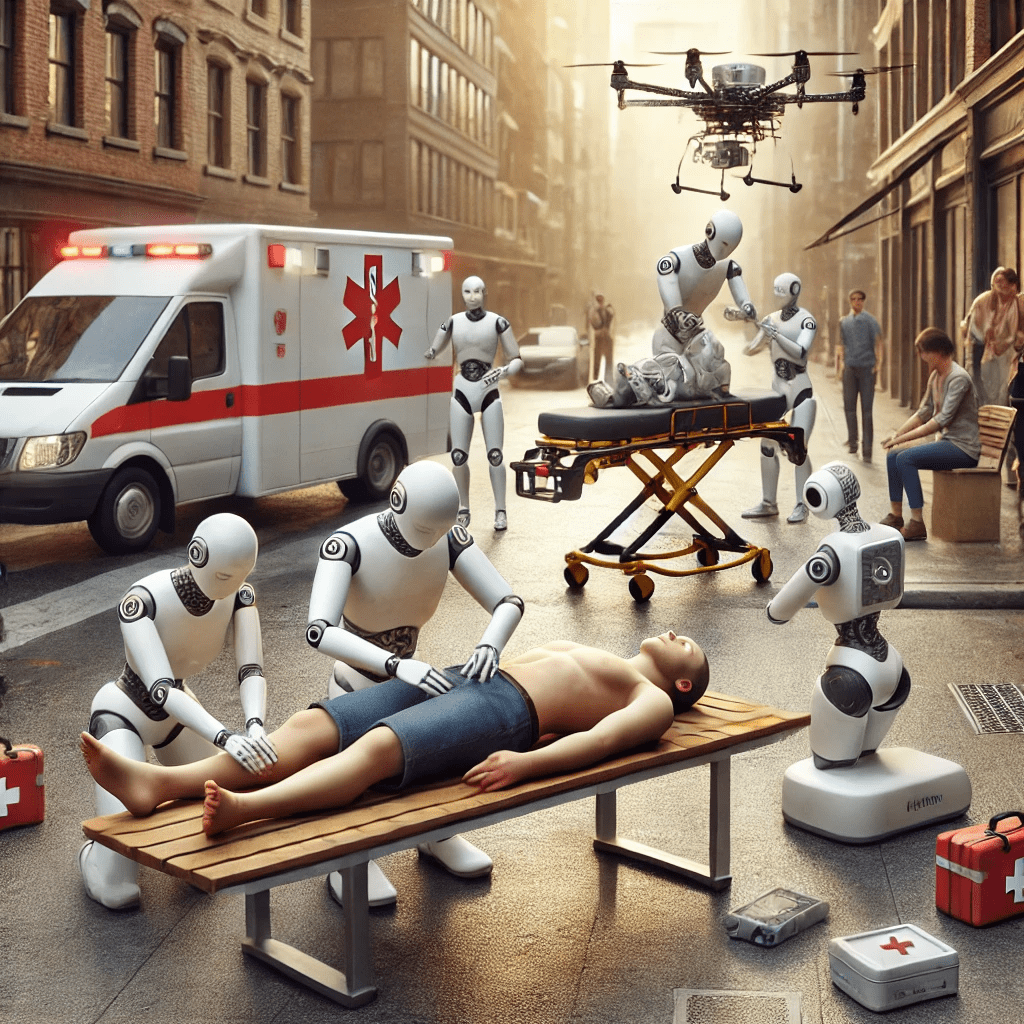}
        \caption{}
        \label{fig:medical}
    \end{subfigure}
    \begin{subfigure}[b]{0.18\textwidth}
	    \centering
        \includegraphics[width=1\linewidth, height=1\linewidth, keepaspectratio]{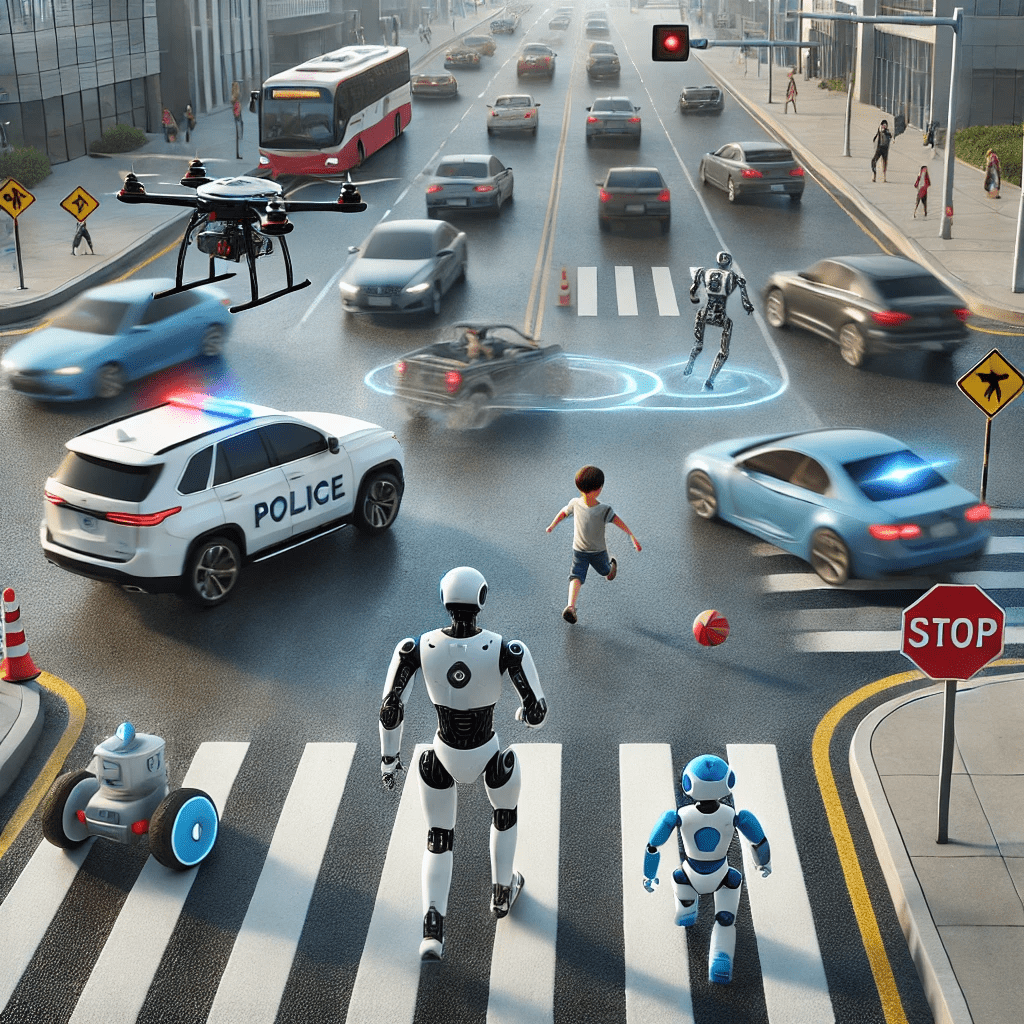}
        \caption{}
        \label{fig:savingChild}
    \end{subfigure}
    
    \caption{\footnotesize{Diverse scenarios involving heterogeneous multi-robot team missions. (a) Assisting an old lady cross the street safely, (b) Search and Rescue, (c) Medical emergency, (b) Ensuring safety of a child chasing a ball. These images were created with the assistance of DALL-E 3.}}
    \label{fig:Scenarios}
\end{figure}

In recent decades, there have been substantial research efforts on the development of cooperative multi-agent systems~\cite{torreno2017cooperative, mahajan2022generalization}. As we look toward the future, we anticipate a society where robots with diverse capabilities become integral to daily life, contributing to a pro-social and harmonious society~\cite{vsabanovic2010robots, paiva2018engineering}. These robots could assist in various ways, such as helping an elderly person navigate urban streets,
performing search and rescue operations, helping in a medical emergency, or ensuring the safety of a child in dangerous situations such as a busy road (see Fig.~\ref{fig:Scenarios}). The heterogeneity of these robots allows them to collaborate and form dynamic teams~\cite{gupta2022incentivizing, gupta2023optimal}, and successfully carry out intricate operations that are required to accomplish complex missions. To realize this potential, it is essential to design generalized mission-planning frameworks that can accommodate the diverse capabilities of heterogeneous multi-robot teams.

Mission and task planning involve breaking down complex objectives into smaller, manageable tasks, ensuring alignment with available resources and adherence to system constraints. In heterogeneous multi-robot teams, each robot, with its own distinct capabilities and limitations, must be strategically assigned tasks that contribute to the mission in a coordinated and efficient manner. While there are many robot- and task-specific planning approaches, ranging from dexterous manipulation~\cite{jakel2012learning, daoud2012real} to trajectory planning for autonomous vehicles~\cite{gupta2023interaction, gupta2022towards,10588713}, the literature lacks generalized mission planning approaches for heterogeneous multi-robot teams. Moreover, existing mission planning approaches for heterogeneous multi-robot teams are often tailored to specific scenarios \cite{roberts2015coordinating}, which limits their flexibility and adaptability across a wide range of real-world applications. 

A key limitation of conventional mission planning systems is their lack of general intelligence and common-sense reasoning, both of which are essential for managing the dynamic and unpredictable nature of real-world environments. Consequently, these systems often struggle to generalize across different mission types, limiting their effectiveness in diverse contexts. Recent advancements in Large Language Models (LLMs) such as GPT~\cite{achiam2023gpt}, Gemini~\cite{team2023gemini}, and LLaMA~\cite{touvron2023llama} have demonstrated emergent capabilities that suggest they might be useful in addressing the shortcomings of mission planning systems to reason over diverse and novel situations. 
Therefore, we hypothesize that incorporating LLMs into mission planning systems enables more robust and adaptable planning for heterogeneous multi-robot teams, significantly increasing their potential for real-world applications.

The mission in Multi-Robot Task Allocation (MRTA) problems can be categorized as elemental, simple, compound, or complex~\cite{korsah2013comprehensive, khamis2015multi}. Among these categories, the most challenging tasks are complex missions with complex task dependencies (CD), where each task can be completed in multiple ways. Our focus is on complex missions of type CD[ST-MR-TA], which involve multiple robots (MR), where each robot performs a single task (ST) at a time. Task allocation and scheduling take into account time-extended assignments (TA), meaning both current and future tasks are considered during planning. 

In recent years, significant progress has been made in multi-robot coordination and task allocation, with researchers exploring various approaches such as probabilistic methods based on partially observable Markov decision processes~\cite{morere2017sequential, gupta2023optimal1, floriano2019planning}, game-theoretic approaches~\cite{bredin2000game, gupta2019achieving}, auction- and market-based strategies~\cite{otte2017multi, nunes2017decentralized}, and optimization-based techniques~\cite{luo2015distributed, sghir2018multi}. Despite their advantages, these methods do not scale and generalize to complex real-world class CD problems, which require task decomposition, allocation, and scheduling under intricate constraints. Moreover, limited studies that are designed for class CD problems, such as those utilizing Generalized Partial Global Planning (GPGP)~\cite{motes2020multi} and hierarchical trees~\cite{9926542}, rely heavily on human expertise for task decomposition and treat task allocation and scheduling as separate processes from task decomposition. In contrast, we leverage the common-sense reasoning and function-calling capabilities of LLMs to construct hierarchical trees, utilizing custom-designed subtree routines and APIs that seamlessly integrate robot information. Our approach offers a more cohesive and fully automated control-loop system that does not require any human intervention.

Our mission-planning strategy involves breaking down the complex mission into executable tasks through a hierarchical tree built using an LLM. The representation of multi-robot missions using hierarchical trees is inspired by the language Task Analysis, Environment Modeling and Simulation (TAEMS)~\cite{decker1996taems, horling1999taems, phelps2005coordinators}. Specifically, it utilizes task decomposition, where intricate tasks are incrementally decomposed into simpler tasks, down to the level of actionable tasks. The hierarchical tree structure offers a clear overview of the mission by outlining task dependencies and streamlining the overall mission design. Furthermore, its flexibility and expressiveness also enable the modeling of complex task relationships, making it highly adaptable across diverse domains. After constructing the hierarchical tree using an LLM, we apply a heuristic tree-search algorithm to decompose the tree into multiple MRTA alternatives, ensuring successful mission accomplishment. 

This work makes two key contributions. First, we present a generalized mission-planning framework based on hierarchical tree construction, leveraging the common-sense reasoning and function-calling capabilities of LLMs. This approach allows the pipeline to handle diverse, complex missions that are challenging for traditional systems. Second, we introduce a heuristic algorithm to decompose the hierarchical tree into multiple MRTA alternatives, ensuring that missions are completed while adhering to system resources and constraints. Consequently, this decomposition facilitates the formation of multi-robot teams, with tasks efficiently assigned to each robot.

\section{Background and Formulation }\label{sec:background}

We present a mission-planning pipeline for a team of heterogeneous robots with diverse constraints and capabilities. This work focuses exclusively on high-level mission planning, assuming the availability of scene understanding~\cite{bremond2007scene} and low-level motion planning~\cite{le2023social, le2024multi} for task execution. 
The mission-planning module receives relevant scene information from the scene-understanding component, interprets the high-level mission objective, formulates a plan, and assigns tasks to the available robots accordingly. Identifying the overall mission objective from scene understanding can be facilitated by an LLM. In this work, we assume the availability of such a mission objective.
Additionally, it is assumed that the low-level planner can successfully execute the  task assignments for each robot. 

Let $\mathcal{R} = \{r_1, \ldots, r_m\} $ be the set of robots. 
Each robot $i \in \mathcal{R}$ can be of different types and can have various capabilities and limitations denoted by the sets $\mathcal{C}_i$ and $\mathcal{L}_i$, respectively. For example, a transportation robot may have the capabilities defined as [reach location, follow agent, carry load] and limitations defined as [max speed = $2$ m/s, max load capacity = $4$ Kg]. Each robot $i$ can perform a set of actions $\mathcal{A}_i$ based on its capabilities $\mathcal{C}_i$. For instance, a robot with the ``follow" capability may have an action set that includes actions such as $\texttt{get\_follow\_path}$, $\texttt{follow\_path}$, and $\texttt{send\_location\_to\_server}$.  
These actions represent various executable sub-routines that can be performed by the robot, and will be referred as primitive actions throughout the paper. The system may have redundancy, and hence, in general $\mathcal{A}_i \intersection \mathcal{A}_j \neq \emptyset$ for $i \neq j, i, j \in \mathcal{R}$. 


\begin{figure}
	\centering
	\includegraphics[width=0.85\linewidth, height=\linewidth, keepaspectratio]{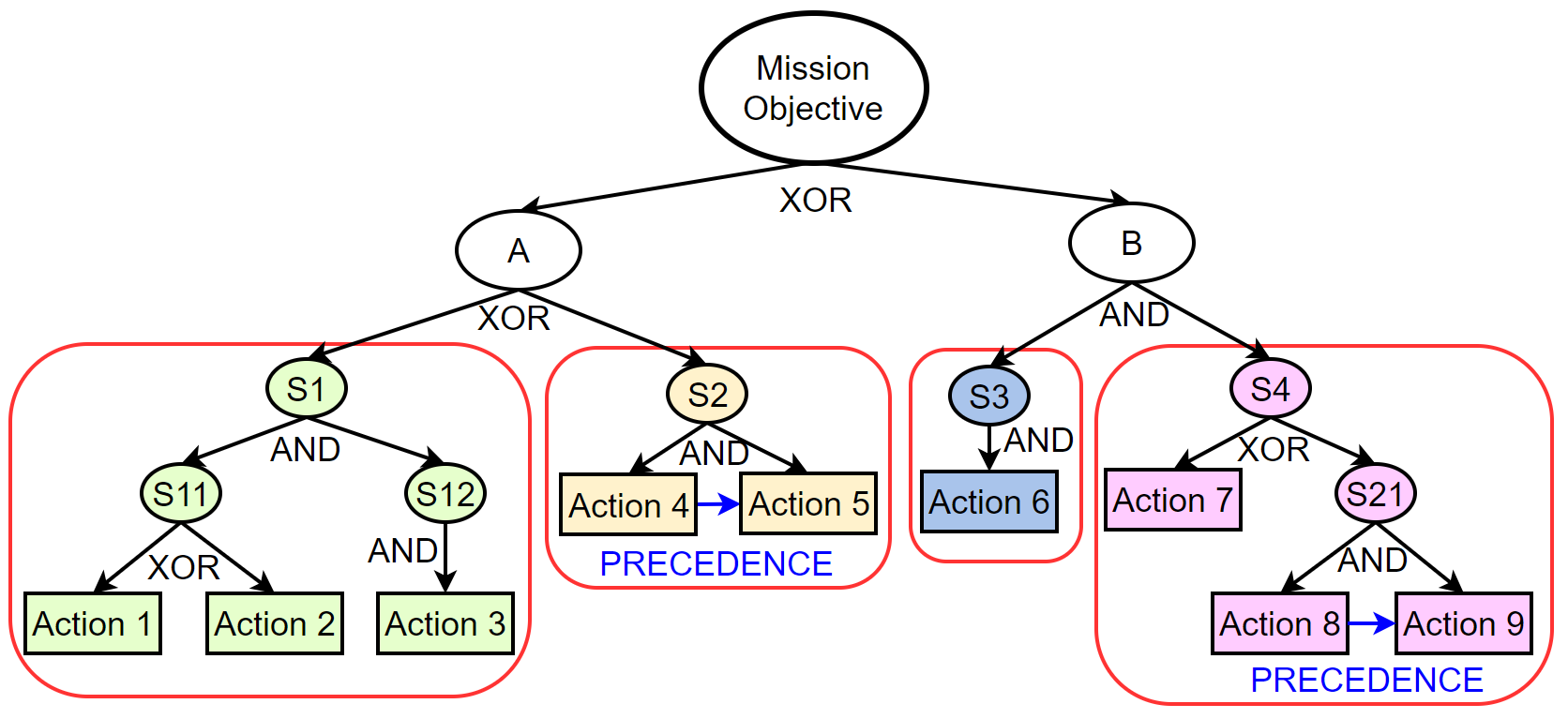}
    \caption{\footnotesize{An example hierarchical tree. Various subtrees of different color are highlighted in red enclosures. The round nodes represent abstract non-primitive tasks and the rectangular nodes represent executable primitive tasks. Children nodes are connected to their parent node with a logical constraint. The blue arrow represents precedence relationship between tasks.
    }} 
    \label{fig:HierarchicalTree}
\end{figure} 

We represent the mission using a hierarchical tree, where the objective is to incrementally decompose the mission objective until the level of primitive actions $a \in \mathcal{A}_i$ that can be executed by the robot $i \in \mathcal{R}$. This approach is inspired by TAEMS, a framework designed to represent large task hierarchies along with the complex relationships between tasks.

Fig.~\ref{fig:HierarchicalTree} provides an example of such a hierarchical tree. The root node represents the overall mission objective, which is progressively decomposed into simpler tasks, represented as nodes in the tree. These task nodes are classified into two types: primitive tasks (depicted as rectangular nodes) and non-primitive tasks (depicted as round nodes). Primitive tasks are the most basic actions a robot can directly execute based on its capabilities, while non-primitive tasks are abstract tasks that combine with primitive tasks to provide a meaningful structure for achieving the mission objective. At each level, child nodes are linked to their parent nodes via a logical constraint $\{AND, XOR\}$. For $AND$, all child tasks must be completed to satisfy the parent task, whereas for $XOR$, exactly one child task must be completed to fulfill the parent task. Additionally, temporal constraints, such as task precedence (illustrated by blue arrows in Fig.~\ref{fig:HierarchicalTree}), may exist between tasks. Collectively, tasks across various levels of the tree contribute to achieving the root-level mission objective. 

To evaluate each task, each primitive task $a \in \mathcal{A}_i$ is assigned a utility $u_a(i)$. In the TAEMS framework~\cite{horling1999taems}, this utility is often defined as a function of a triple $(q_a(i), d_a(i), c_a(i))$, where $q_a(i)$ represents the quality of the action, $d_a(i)$ represents the time duration of the action, and $c_a(i)$ estimates the cost for the action (energy expenditure, financial cost, resource consumption etc.) performed by the robot $i$. These functions are designed by the system designer and can be estimated by the robot based on its current state. The utility of the action $a$ taken by robot $i$ can be computed as:
\begin{equation}\label{eq:primitive_utility}
    u_a(i) = \alpha q_a(i) -\beta d_a(i) - \gamma c_a(i),
\end{equation}
where $\alpha,\beta, \gamma$ are the system hyper-parameters.

\section{Mission Planning Framework}\label{sec:framework}

We now discuss the mission-planning framework consisting of hierarchical tree construction using LLMs and tree decomposition into MRTA alternatives.

\subsection{Hierarchical Tree construction via LLM function calling}

The goal of the mission planning is to map the mission objective into a set of multi-robot task assignments, consisting of primitive tasks that can be executed by the robots. This becomes particularly challenging in complex class CD problems, where the mission can be accomplished in multiple ways, and tasks may have intricate dependencies. To address this, we leverage the intelligence and common-sense reasoning of LLMs to construct hierarchical trees for a wide range of real-world complex missions.

However, while LLMs are strong in common-sense reasoning, their outputs can be unstructured and prone to hallucinations. This challenge makes it difficult for LLMs to construct a well-organized hierarchical tree with zero-shot prompting. To overcome this, we design various tools and APIs to assist the LLM, utilizing its function-calling capabilities to reliably generate the hierarchical tree.

\begin{figure}[!ht]
	\centering
	\includegraphics[width=0.85\linewidth, height=\linewidth, keepaspectratio]{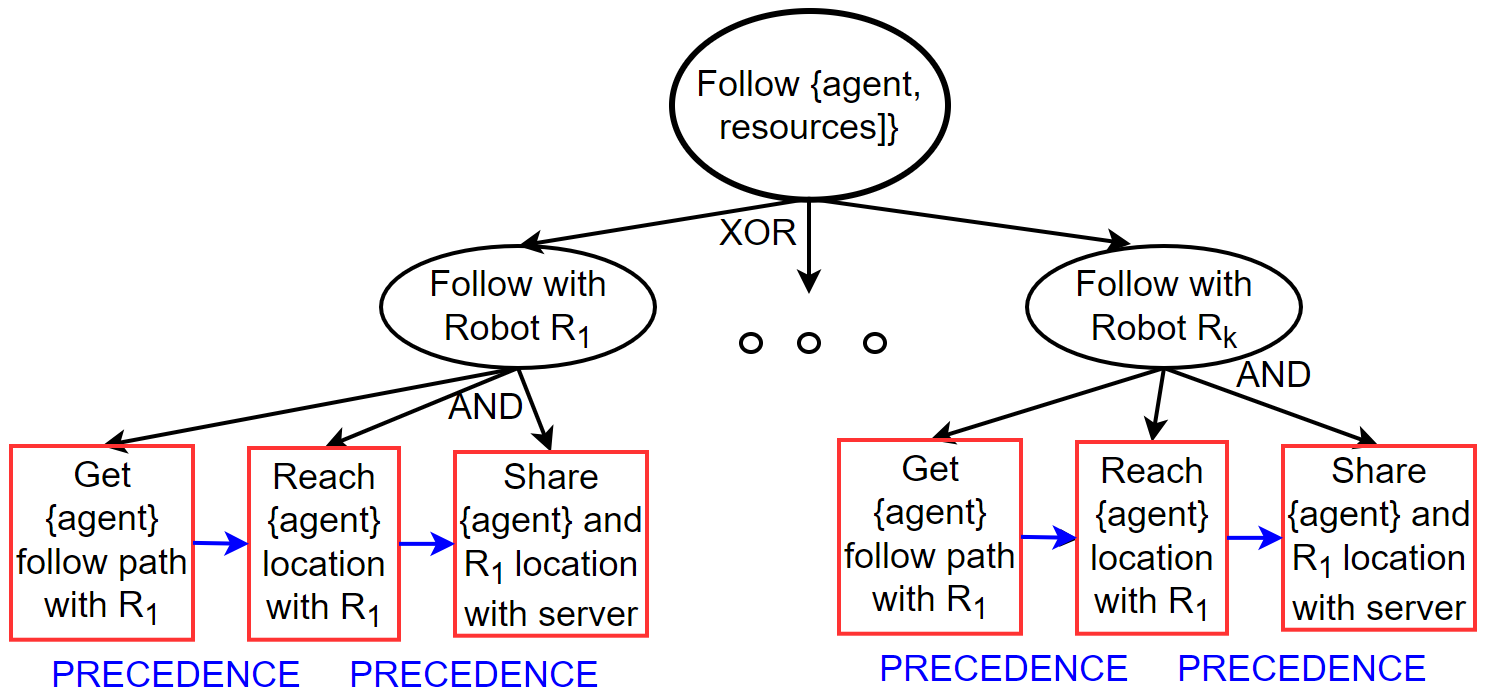}
    \caption{\footnotesize{Follow Subtree constructed by the \texttt{FollowSubtree (agent, resources)} routine. Robots that have follow capability in the resources are included in the subtree. The red rectangular nodes represent the primitive actions available to the robots with follow capabilities.
    }} 
    \label{fig:FollowSubtree}
\end{figure}

In the hierarchical tree, all leaf nodes terminate at primitive tasks, which are defined based on the capabilities of the robots. Given a set of robot resources $\mathcal{R}$, we construct a predefined set of subtree routines corresponding to system capabilities, denoted by $\mathcal{C} = \mathcal{C}_1 \union \ldots \mathcal{C}_m$. Each subtree routine is a function that creates a pre-defined subtree based on system resources, constraints, and other functional arguments, e.g.,  \texttt{FollowSubtree (agent, resources)} routine constructs a follow subtree (Fig.~\ref{fig:FollowSubtree}) to follow an agent while adhering to robot capabilities and constraints. 

This approach offers multiple advantages. First, for each subtree, we can include a node for all (or a subset of) robots capable of performing a subtree task, thereby integrating the multi-robot team formation (robot selection) process directly into the hierarchical tree construction. Second, multiple subtrees can be merged together to create more complex subtree routines, such as $\texttt{SearchAndFollowSubtree (agent, resources)}$, to search and follow an agent. Lastly, since all subtrees adhere to the hierarchical structure and are common across a wide range of missions (as they are often context-independent), the LLM can leverage these predefined subtree routines to automatically generate the necessary subtrees and seamlessly integrate them into the overall hierarchical tree. Similar to primitive tasks, all hierarchical trees constructed via LLM should now terminate with a subtree attachment.

\begin{figure}[!ht]
	\centering
	\includegraphics[width=\linewidth, height=\linewidth, keepaspectratio]{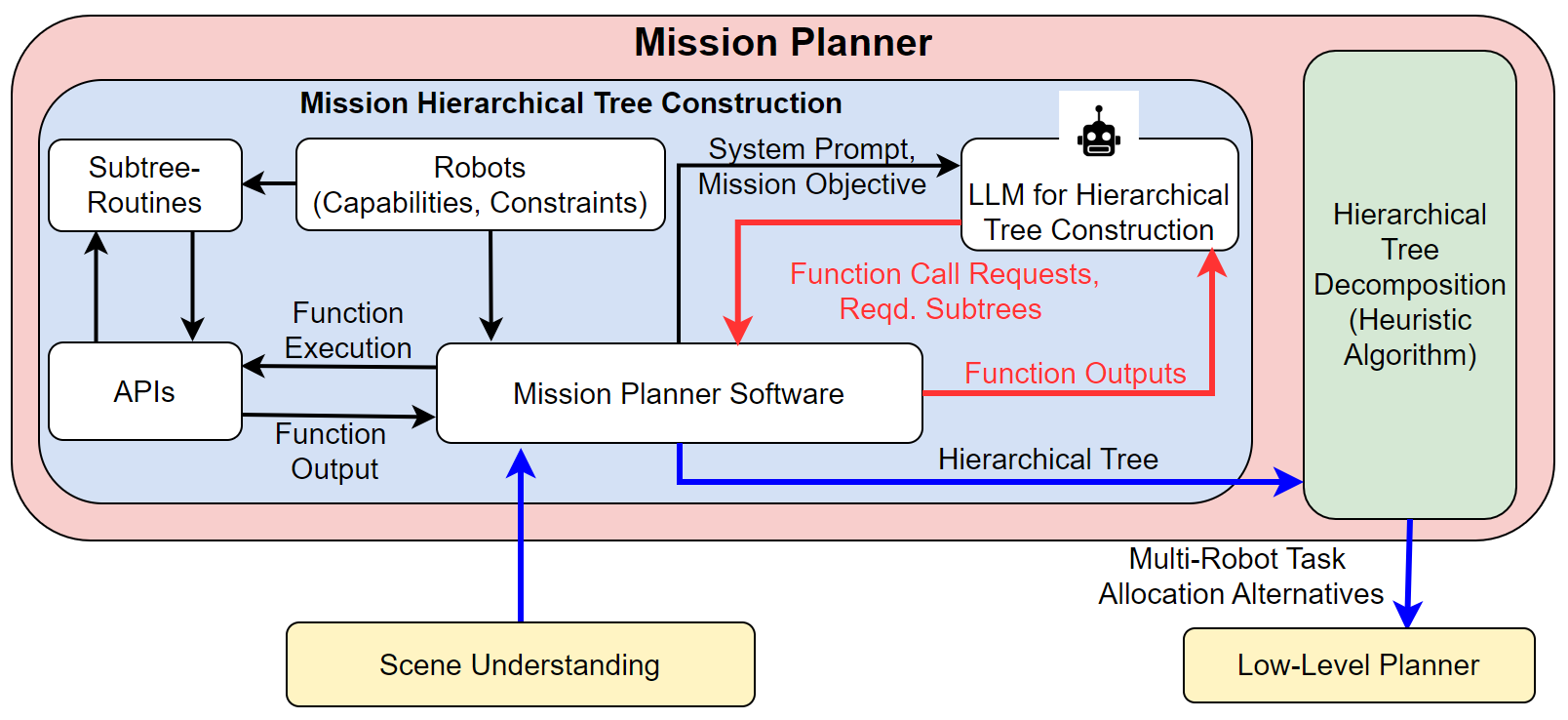}
    \caption{\footnotesize{Overall mission planning pipeline. LLM utilizes the subtree-routines and APIs to create a hierarchical tree which is decomposed to obtain MTRA alternatives.
    }} 
    \label{fig:Pipleline}
\end{figure}

In addition to the subtree routines, we design a set of APIs such as \texttt{HierarchicalTreeConstructor}, \texttt{CreateAndAddSubtask}, \texttt{addMultiSubtasks}, \texttt{attachMultiSubTrees}, \texttt{plotTree}, \texttt{printTree}, and others to assist the LLM in constructing the hierarchical tree. The function definitions, along with details of their arguments, are provided to the LLM in JSON format. The LLM then leverages its function-calling ability to request these function calls with the appropriate arguments. The system executes the function calls requested by the LLM, ensuring the reliable construction of the hierarchical tree while reducing the LLM's overhead. Fig.~\ref{fig:Pipleline} shows the overall mission planning pipeline. 

\subsection{Generating MRTA Alternatives}

\begin{algorithm}[hbt!]
\small
\caption{MRTA Alternatives}\label{alg:task_alternatives}
\textbf{Parameters:} $\rho$ - Maximum number of alternatives \\
\textbf{Input:} Hierarchical tree $\mathcal{T}$, available resources $\mathcal{R}$, task \textit{task} \\
\textbf{Output:} Alternative decomposition of $\textit{task}$, consumed resources \\

\SetKwFunction{FMain}{get\_alternatives}
\SetKwProg{Fn}{Function}{:}{}
\Fn{\FMain{task}}{
\texttt{\blue{/* recursion stopping criterion */}} \\
    \If{$task \in \mathcal{A}$}{ 
    $resource\_used \gets \texttt{check\_consumption}(task, \mathcal{R})$ \\
        \textbf{return} [[[$task$]]], \ [[$resource\_used$]] 
    }
\texttt{\blue{/* generate alternatives */}} \\
$sub\_list, sub\_resource \gets []$ \\
 \For{$subtask \in \mathcal{T}.subtasks(task)$}{ 
        $temp\_subtask, temp\_res \gets \texttt{get\_alternatives}(subtask)$ \\
      $sub\_list \gets sub\_list \union temp\_subtask $ \\
$sub\_resource \gets sub\_resource \union temp\_res$ \\  
    }
    $task\_alt$ $\gets []$ \\
\Switch{$\mathcal{T}$.constraints(task)}{
\Case{AND}{
$temp\_alt \gets \texttt{cartesian\_product}(alt)$ \\
$temp\_resource \gets \texttt{update\_resource} (sub\_resource,  temp\_alt)$ \\
\If{not \texttt{resource\_fail}$(temp\_resource, \mathcal{R})$}{
    $task\_alt \gets temp\_alt$ \\
    $task\_resource \gets temp\_resource$
}
\Else{
$task\_alt \gets [[]]$ \\
    $task\_resource \gets []$
}

}
\Case{XOR}{
$task\_alt \gets [[alt^*] \ for \ alt^* \in alt]$ \\
$task\_resource \gets [[r^*] \ for \ r^* \in sub\_resource]$
}
}
\texttt{\blue{/* pruning */}} \\
\If{$|task\_alt| > \rho$}{
       $utility \gets \texttt{compute\_utility} (task\_alt)$\\
       $top\_ix \gets \texttt{top\_sorted\_indices} (utility, \rho)$ \\
       $task\_alt \gets [task\_alt(i), i \in top\_ix]$ \\
       $task\_resource \gets [task\_resource(i), i \in top\_ix]$ \\
}
return $task\_alt, \ task\_resource$
}
\end{algorithm}


To decompose the hierarchical tree into several MRTA alternatives, we develop a heuristic algorithm as outlined in Algorithm~\ref{alg:task_alternatives}. This recursive algorithm begins at the primitive nodes and progressively builds upwards, concluding at the root node. Let $alt(task) \subseteq \mathcal{A} = \mathcal{A}_1 \union \ldots \mathcal{A}_m$ represent an unordered set of all actions that, when executed, lead to the completion of $task$.

The cardinality of the $alt$ set is influenced by factors such as the mission tree, available resources, and constraints. For missions with abundant resources and minimal constraints, the MRTA alternative generation process can experience combinatorial explosion, resulting in factorial complexity. To address the potential combinatorial explosion, the heuristic algorithm prunes the minimum-utility results at each step to make the problem tractable. During the procedure, the algorithm computes the utility of the set  $task\_alt$ by summing the utility of each primitive task in the set, i.e., $u(task\_alt) = \sum_{ a(i) \in task\_alt} u_a(i)$. It is important to note that, while the greedy approach in the pruning stage leads to a computationally efficient algorithm, it can produce sub-optimal solutions, particularly in resource-constrained systems. However, in most real-world missions, computational efficiency is often preferred over achieving globally optimal allocations.


Once MRTA alternatives are obtained from Algorithm~\ref{alg:task_alternatives}, we utilize a topological sort algorithm~\cite{cormen2022introduction} to satisfy the precedence constraints in each robots task allocation. 


\section{Results }\label{sec:results}
We utilize the state-of-the-art GPT-4o-2024-08-06 as our chosen LLM for this work, due to its capability to generate structured outputs and reliably execute function calls. We employ a system message and chain-of-thought prompting~\cite{wei2022chain, wang2022self} to obtain reliable hierarchical trees. In our experiments, we focus on mobility tasks and select a set of robots as outlined in Table~\ref{tab:robots}. Each robot has its set of capabilities, constraints, and primitive actions. For brevity, we omit the detailed descriptions. For all primitive actions, instead of defining utility for each action individually, we assign a uniform utility based on the robot’s capability. Specifically, each primitive action of robot $i$ in a subtree corresponding to a capability receives the same utility, as shown in Table~\ref{tab:robots}. We use a simple utility function $u_a(i) = q_a(i)$ and set $d_a(i), c_a(i)=0$.
However, the framework allows for complex utility functions based on the system designer's preferences.  The following subtree routines:
\begin{enumerate}
    \item \texttt{Search (agent,  resources)}
    \item \texttt{Follow (agent, resources)}
    \item \texttt{Reach (agent, resources)}
    \item  \texttt{SearchAndFollow (agent, resources)}
    \item \texttt{Transport (agent, location)}
    \item \texttt{ReachAndTransport (agent, location)}
\end{enumerate}
and following function APIs:
\begin{enumerate}
    \item  \texttt{HierarchicalTree\_init (objective)}
    \item  \texttt{CreateAndAddSubtask (parentString, isPrimitive, taskName, LogicalConstraint)}
    \item  \texttt{addMultiSubtasks (parentString, isPrimitiveList, taskNameList, LogicalConstraint)}
    \item  \texttt{attachMultiSubTrees (parentString,  treeNames, treeArguments, LogicalConstraint, contraintsPairs)}
    \item  \texttt{plotTree ()}
    \item \texttt{printTree ()}
\end{enumerate}
were provided to the LLM.

\begin{table}[!ht]
\centering
\begin{tabular}{|c|c|c|}
\hline
\textbf{Robot Type} &
  \textbf{\begin{tabular}[c]{@{}c@{}}Num. of \\ robots\end{tabular}} &
  \textbf{\begin{tabular}[c]{@{}c@{}}Capabilities \& \\ Primitive Action Utilities\end{tabular}} \\ \hline
Mobile Scooter & 1 & Follow: 0.5, Reach: 1.0                                                                  \\ \hline
Tele-Robot     & 2 & \begin{tabular}[c]{@{}c@{}}Search: 1.0, Follow: 0.5, \\ Reach: 1.0\end{tabular}          \\ \hline
\begin{tabular}[c]{@{}c@{}}Transportation\\  Robot\end{tabular} &
  1 &
  \begin{tabular}[c]{@{}c@{}}Search: 0.3, Follow: 0.3,\\  Reach: 0.3, Carry: 1.0\end{tabular} \\ \hline
Social Robot   & 1 & \begin{tabular}[c]{@{}c@{}}Follow: 0.3, Reach: 0.3, \\ Message Display: 1.0\end{tabular} \\ \hline
\end{tabular}
\caption{Resources}
\label{tab:robots}
\end{table}

\begin{figure*}[ht]
    \centering
	\begin{subfigure}[b]{0.31\textwidth}
	    \centering
        \includegraphics[width=1\linewidth, height=1\linewidth, keepaspectratio]{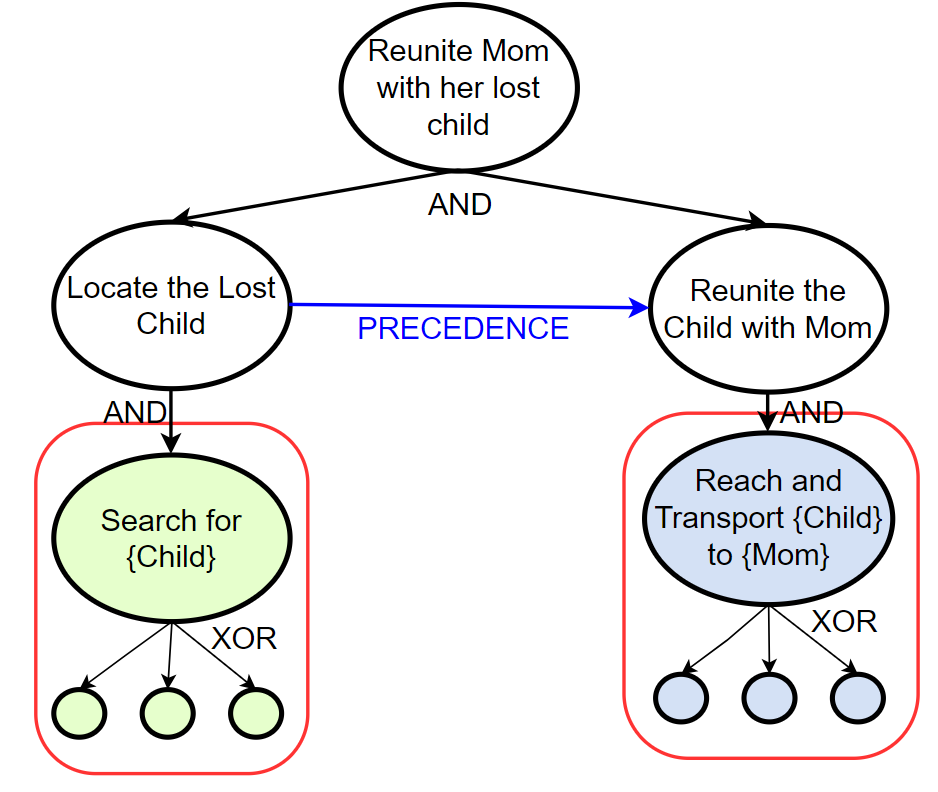}
        \caption{}
        \label{fig:momChild}
    \end{subfigure}
    ~
    \begin{subfigure}[b]{0.31\textwidth}
	    \centering
        \includegraphics[width=1\linewidth, height=1\linewidth, keepaspectratio]{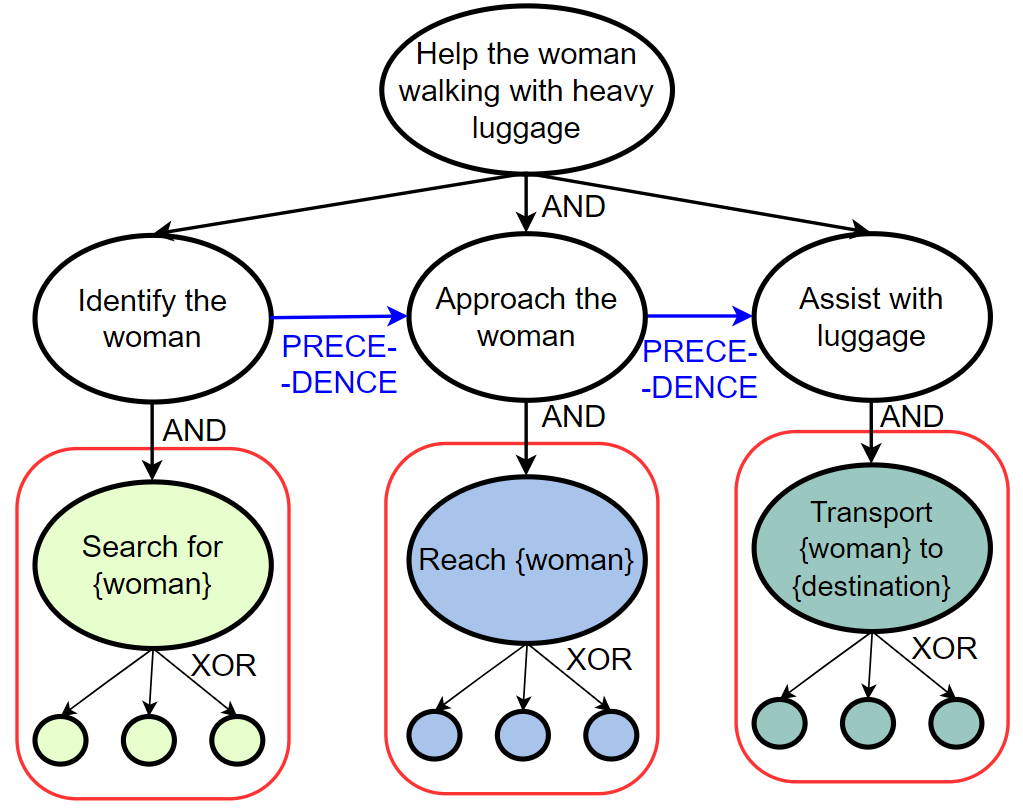}
        \caption{}
        \label{fig:luggage}
    \end{subfigure}
    ~
    	\begin{subfigure}[b]{0.31\textwidth}
	    \centering
        \includegraphics[width=1\linewidth, height=1\linewidth, keepaspectratio]{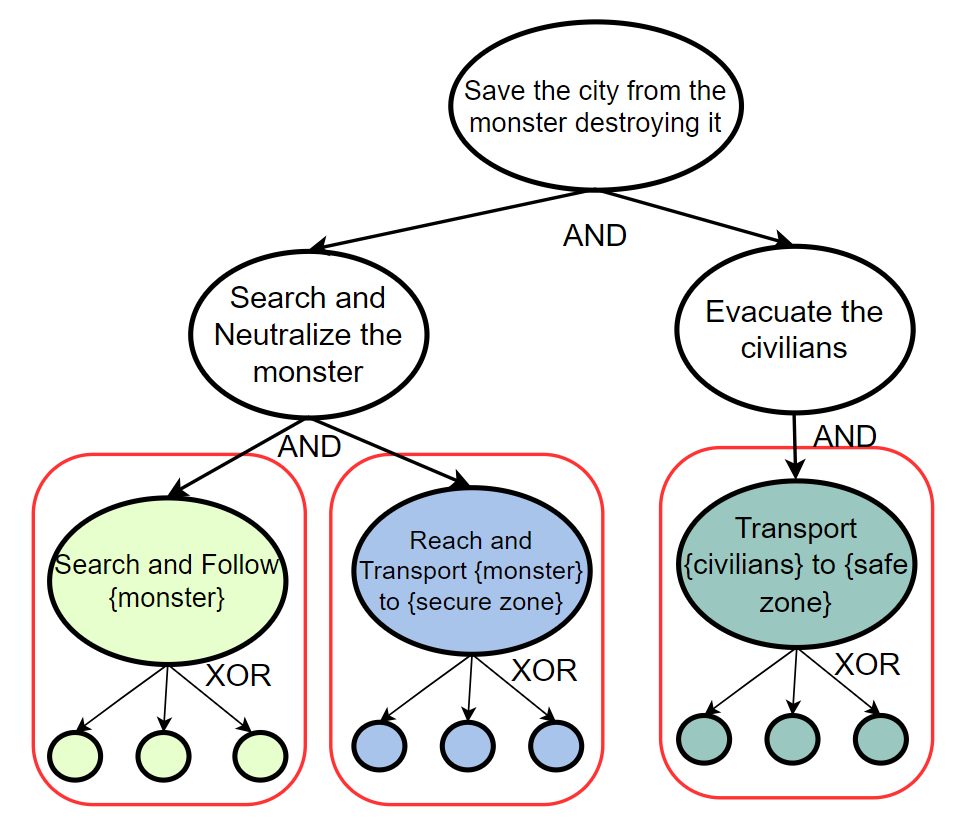}
        \caption{}
        \label{fig:saveCity}
    \end{subfigure}
        ~
    	\begin{subfigure}[b]{0.31\textwidth}
	    \centering
        \includegraphics[width=1\linewidth, height=1\linewidth, keepaspectratio]{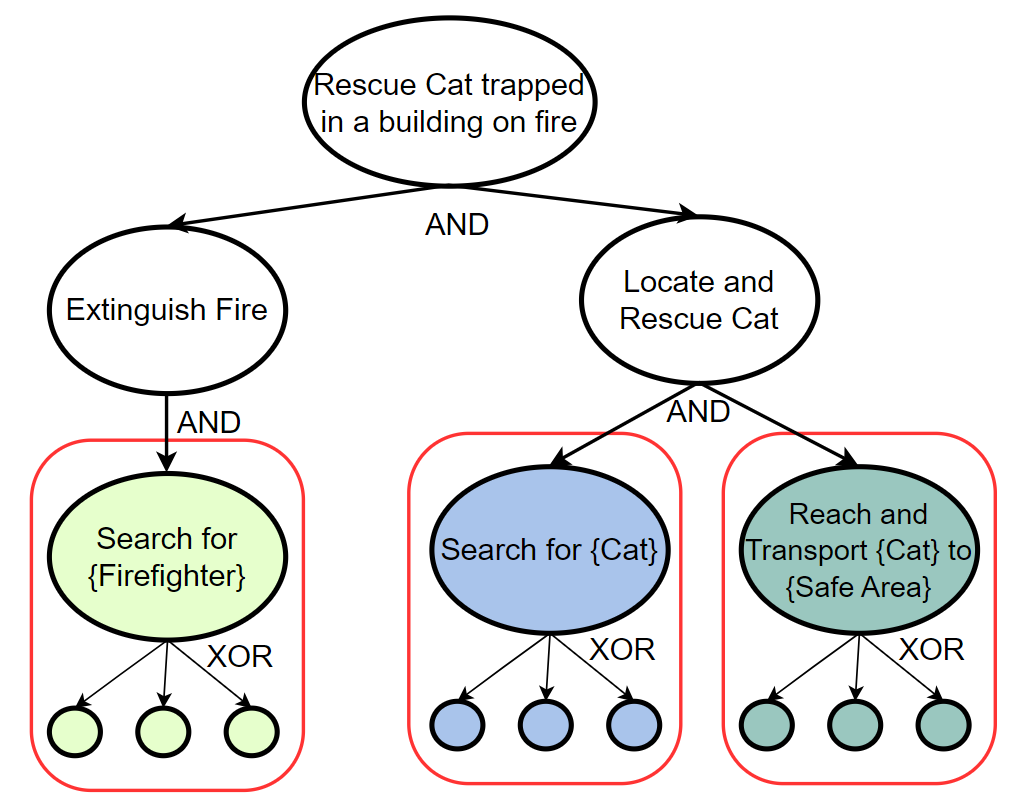}
        \caption{}
        \label{fig:rescueCat}
    \end{subfigure}
            ~ ~ ~
    	\begin{subfigure}[b]{0.44\textwidth}
	    \centering
        \includegraphics[width=1\linewidth, height=1\linewidth, keepaspectratio]{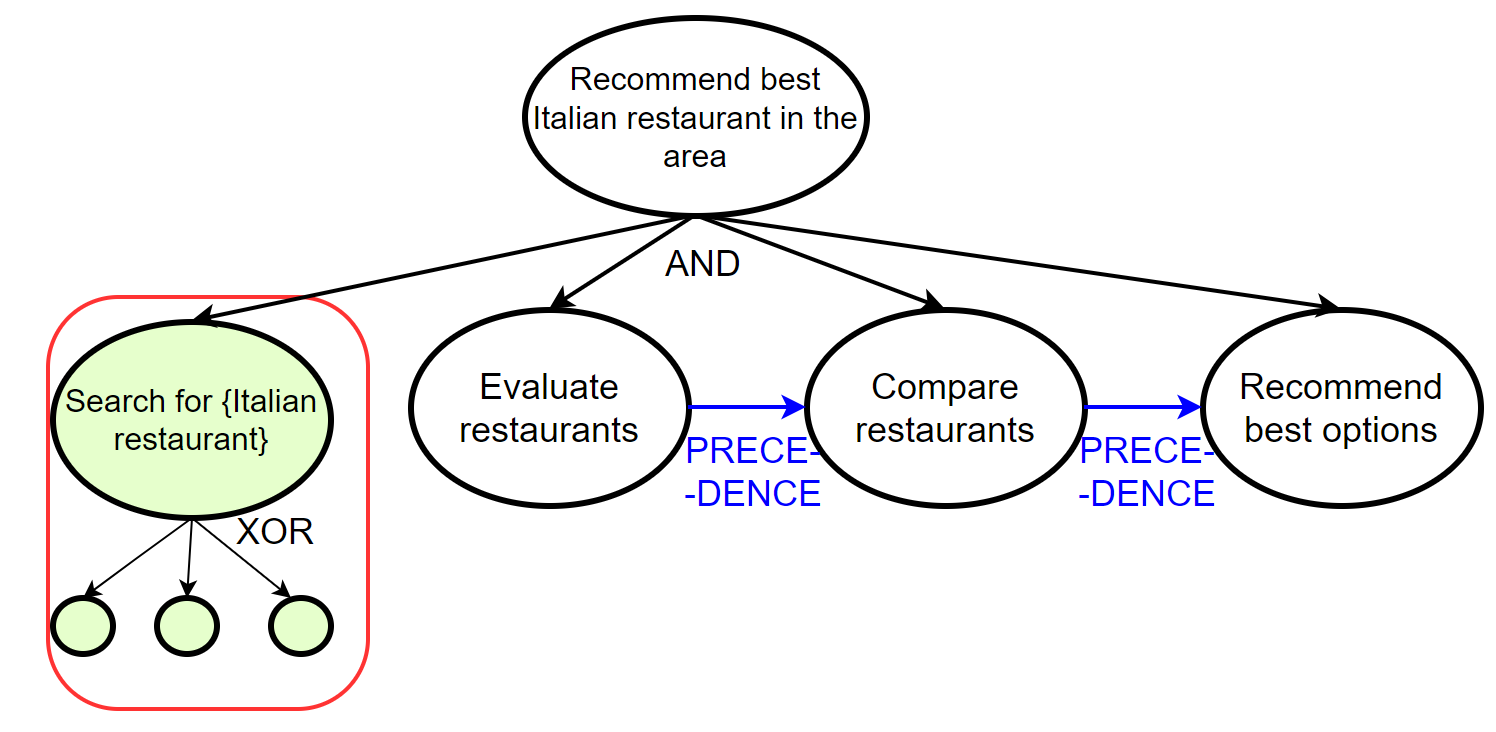}
        \caption{}
        \label{fig:restaurants}
    \end{subfigure}
    \caption{\footnotesize{Hierarchical trees  generated by the LLM for different mission goals: (a) Reunite mom with her lost child, (b) Help the woman walking with heavy luggage, (c) Save the city from the monster destroying it, (d) Rescue cat trapped in a building on fire, (e) Recommend best Italian restaurants in the area. The details of the colored subtrees (highlighted with red boxes) used to create the hierarchical tree are omitted for brevity of space. 
    }}
    \label{fig:LLMtrees}
\end{figure*}

The following is an example of a hierarchical tree generation process by the LLM for the mission objective "Reunite mom with her lost child" (see also Fig.~\ref{fig:momChild}). 

\begin{mdframed}[style=custombox]
\small
\red{System:} You are an advanced multi-agent decision-making system responsible for generating a Hierarchical Tree for a given mission objective. The process begins by establishing the mission objective as the root node. From this root, you can add nodes or entire subtrees to build the structure. Each node represents a subtask derived from its parent task, and the children of each node are connected through logical constraints that define their relationships. The hierarchical structure must ensure that all paths terminate in a valid subtree, meaning that every leaf node must be a subtree itself. You have access to a variety of predefined subtrees, which you can select based on the task requirements with appropriate arguments. To assist in creating the optimal tree, you can utilize various functions available through tool calls. After each tool call, you will receive the output from the corresponding function, which you can use to refine your decision-making process and continue building the tree. Your goal is to construct a coherent and logically consistent tree that effectively represents the mission's objectives and constraints. Before constructing the hierarchical tree by calling functions, think step by step how would you construct this tree. Think about the important agents, available subtree routines, and arguments you would use to create these subtrees. Remember that in the hierarchical tree, each branch must terminate with a subtree. Note that a hierarchical tree can contain multiple ways of solving the mission objective through the use of appropriate logical constraints. 

\noindent\rule{\textwidth}{1pt}

\blue{User:} Create a Hierarchical Tree with the mission - `Reunite mom with her lost child'. Think how to use the functions and subTrees to create a logical Hierarchical tree.

\noindent\rule{\textwidth}{1pt}
\red{assistant:} 
To create a Hierarchical Tree for the mission objective ``Reunite mom with her lost child", we will need to identify the key agents involved and utilize the available sub-trees in a manner that creates a coherent and optimal decision-making structure. \\
\noindent\orange{\#\#\# Key Agents:} \\
1. **Mom** - We need to find the child and potentially transport the child to mom. \\
2. **Child** - The child is lost, and our primary focus is to reunite them with the mom.\\
\noindent\orange{\#\#\# Available Subtrees:}\\
1. **SearchTree(agent)** - To search for a specific agent.\\
2. **FollowTree(agent)** - To follow a specific agent.\\
3. **ReachTree(agent)** - To reach a specific agent.\\
4. **SearchAndFollowTree(agent)** - To search for and then follow an agent. \\
5. **TransportTree(agent, destination)** - To transport an agent to a destination. \\
6. **ReachAndTransportTree(agent, destination)** - To reach an agent and then transport them to a destination. \\
\noindent\orange{\#\#\# Hierarchical Tree Construction:} \\
We can approach this problem by breaking it into logical subtasks: \\
1. **Locate the Lost Child** - We first need to find the child. \\
2. **Transport the Lost Child to Mom** - Once the child is found, we need to transport the child back to mom. \\
\noindent \orange{\#\#\#\# Steps:} \\
1. **Locate the Lost Child**:
   - **SearchTree** for the child. \\
2. **Reunite the Child with Mom**: 
   - Once the child is located, use the **ReachAndTransportTree** to reach the child and then transport them to the mom. \\
\noindent Now, let's proceed to use the available functions to create this hierarchical tree. We'll start by initializing the tree and then attaching the respective subtasks and subtrees. \\
\noindent\rule{\textwidth}{1pt}
\red{assistant:}  [{`\orange{function}': {`name': `HierarchicalTree\_init', `arguments': \{``objective": ``Reunite mom with her lost child"\}}},  \ \ \ \
{`\orange{function}': {`name': `addMultiSubtasks', `arguments': `\{``parentString": ``Reunite mom with her lost child", ``isPrimitiveList": [false, false], ``taskNameList": [``Locate the Lost Child", Reunite the child and mom], ``LogicalConstraint": ``AND", ``constraintPairs: " [[``Locate the Lost Child", Reunite the child and mom, Precedence]] \}'}}

\noindent\rule{\textwidth}{1pt}
\begin{flushright}
\blue{function:} (HierarchicalTree\_init): HierarchicalTree constructor was successfully called. \\
\blue{function:} (addMultiSubtasks): was called.
\end{flushright}

\noindent\rule{\textwidth}{1pt}
\red{assistant:} [{`\orange{function}': {`name': `attachMultiSubTrees', `arguments': `\{``parentString": ``Locate the Lost Child", ``treeNames": [``SearchTree"], ``treeArguments": [[``Child"]], ``LogicalConstraint": ``AND", ``contraintsPairs": \{\}\}'}}  \ {`\orange{function}': {`name': `attachMultiSubTrees', `arguments': `\{``parentString": ``Reunite the Child with Mom", ``treeNames": [``ReachAndTransportTree"], ``treeArguments": [[``Child", ``Mom"]], ``LogicalConstraint": ``AND", ``contraintsPairs": \{\}\}'}} ]

\noindent\rule{\textwidth}{1pt}
\begin{flushright}
\blue{function:} (attachMultiSubTrees): was called \\
\blue{function:} (attachMultiSubTrees): was called
\end{flushright}

\noindent\rule{\textwidth}{1pt}
\red{assistant:} [{`\orange{function}': {`name': `plotTree', `arguments': `\{\}'}}]

\noindent\rule{\textwidth}{1pt}
\begin{flushright}
\blue{function:} (plotTree): was called
\end{flushright}

\noindent\rule{\textwidth}{1pt}
\red{assistant:}
The hierarchical tree for the mission "\``Reunite mom with her lost child" has been constructed and plotted successfully.
\end{mdframed}

For the mission - ``Reunite mom with her lost child", the top two task assignments generated by running the heuristic MRTA algorithm are as follows:

\begin{enumerate}

     \item[A1:]  \begin{itemize}
        \item Tele-robot : [Get search path to find {Child}, Detect {Child}, Report detection of {Child} to server]
        \item Social robot : [Get {Child} location to reach, Get to {Child} location, Display message to {Child} to follow Social robot to {Mom}, Get to {Mom} location]
    \end{itemize}
    
    \item[A2:]  \begin{itemize}
        \item Tele-robot : [Get search path to find {Child}, Detect {Child}, Report detection of {Child} to server]
        \item Transportation robot : [Get {Child} location to reach, Get to {Child} location, Carry {Child}, Get to {Mom} location]
    \end{itemize}
\end{enumerate}


The pipeline generates feasible MTRA alternatives to achieve the mission objective. It is important to note that, in practice, when integrated with a low-level planner, tasks allocated to each robot must be scheduled according to the inter-robot task precedence constraints. For instance, in alternative A1, the tasks assigned to the social robot should be scheduled only after the tele-robot has successfully completed its assigned tasks. Figs.~\ref{fig:luggage},~\ref{fig:saveCity}, and~\ref{fig:rescueCat} show more successful hierarchical trees constructed by the LLM for the missions - ``Help the woman walking with heavy luggage", ``Save the city from the monster destroying it", and ``Rescue cat trapped in a building on fire", respectively. For brevity, we omit the chain-of-thought reasoning and function calls made by the LLM. The mission planning pipeline creates successful MTRA alternatives corresponding to hierarchical trees shown in~Figs.~\ref{fig:luggage} and~\ref{fig:saveCity}.   
However, for the mission: ``Rescue cat trapped in a building on fire", the top two task assignments generated by running the heuristic MRTA algorithm are as follows:

\begin{enumerate}

     \item[A1:]  \begin{itemize}
        \item Tele-robot 1 : [Get search path to find {firefighter}, Detect {firefighter}, Report detection of {Firefighter} to server]
        \item Tele-robot 2 : [Get search path to find {cat}, Detect {cat}, Report detection of {cat} to server]
   \item Transportation robot : [Get {cat} location to reach, Get to {cat} location, Carry {cat}, Get to {safe area} location]
    \end{itemize}
    
    \item[A2:]  \begin{itemize}
        \item Tele-robot 1 : [Get search path to find {firefighter}, Detect {firefighter}, Report detection of {Firefighter} to server]
        \item Tele-robot 2 : [Get search path to find {cat}, Detect {cat}, Report detection of {cat} to server]
   \item Social robot : [Get {cat} location to reach, Get to {cat} location, Display message to {cat} to follow social robot to {safe area}, Get to {safe area} location]
    \end{itemize}
\end{enumerate}

In this example, while the first alternative offers a good choice, the second alternative involves a social robot using the primitive action ``Display message to cat to follow social robot to safe area," which may not be feasible in the real world. System designers should account for such edge cases when designing subtree routines. These cases can be avoided in several ways, such as by restricting the types of arguments allowed in subroutines or by selecting a different alternative altogether.

Lastly, Fig.~\ref{fig:restaurants} shows the incomplete hierarchical tree constructed by the LLM for the mission - ``Recommend best Italian restaurants in the area". While the LLM can effectively decompose the mission, it fails to terminate the tree with subtrees that contain the robots' primitive actions. In such cases, where the mission falls outside the scope of the robots' available capabilities and predefined subtree routines, the LLM is unable to generate a complete hierarchical tree.

These examples demonstrate how our framework can leverage the same subtree routines and APIs to produce accurate mission plans for a diverse range of complex missions. 

\section{Limitations and Future Work}\label{sec:Limitations}

The system has a few limitations. In missions where available robot capabilities are insufficient, the LLM may hallucinate, leading to unrealistic solutions. Naturally, without adequate resources, not all problems are feasible. The performance of the system relies on the available subtree routines. Therefore, it is expected
from the system designer to design a comprehensive set of subtree routines that align with the robot capabilities.
Another limitation lies in the heuristic tree-search algorithm, which, due to its pruning process, may not always yield globally optimal solutions. Investigating alternative MRTA algorithms with provable optimality properties is an important avenue for future research. Additionally, the system's overall performance is closely tied to the reliability and efficiency of the scene-understanding module and low-level task execution.

In future work, we aim to conduct closed-loop simulations that integrate the scene-understanding and low-level planner modules. Additionally, we would like to fine-tune~\cite{dodge2020fine} the LLM for hierarchical tree construction task to obtain better quality trees. Lastly, Developing re-planning strategies for mission failures will also be a key focus.

\section{Conclusions}\label{sec:Conclusions}
In this paper, we have introduced a mission-planning framework specifically designed for heterogeneous multi-robot teams. The framework utilizes hierarchical trees to represent and systematically decompose complex missions into manageable tasks. We developed various APIs and tools to automate the construction of these hierarchical trees, leveraging the function-calling capabilities of large language models (LLMs). We present a heuristic algorithm that provides multiple alternative multi-robot task assignments to accomplish the mission objective. Our results demonstrate that the framework effectively generates optimized schedules for the robots, ensuring that each task aligns with the unique constraints and capabilities of eachindividual robot.


\bibliographystyle{IEEEtran}
\bibliography{ref}

\end{document}